\documentclass[10pt,twocolumn,letterpaper]{article}

\usepackage[pagenumbers]{cvpr} 

\usepackage{graphicx}
\usepackage{amsmath}
\usepackage{amssymb}
\usepackage{booktabs}

\usepackage[pagebackref,breaklinks,colorlinks]{hyperref}

\usepackage[capitalize]{cleveref}
\crefname{section}{Sec.}{Secs.}
\Crefname{section}{Section}{Sections}
\Crefname{table}{Table}{Tables}
\crefname{table}{Tab.}{Tabs.}

\def\cvprPaperID{9278} 
\def\confName{CVPR}
\def\confYear{2024}

\newcommand{\my}[1]{\textcolor{black}{#1}}
\begin{document}

\title{MetaDreamer: Efficient Text-to-3D Creation With Disentangling Geometry and Texture}


\author{
Lincong Feng$^{12}$\footnotemark[1]\hspace{0.5cm}
Muyu Wang$^{13}$\footnotemark[1]\hspace{0.5cm}
Maoyu Wang$^{1}$\hspace{0.5cm}
Kuo Xu$^{14}$\hspace{0.5cm}
Xiaoli Liu$^{1}$\footnotemark[2]\hspace{0.5cm}\\
${}^{1}$MetaApp AI Research\hspace{1.5cm}
${}^{2}$Beijing University Of Technology\hspace{1.5cm} \\
${}^{3}$Beijing Institute of Technology\hspace{1.5cm}
${}^{4}$Zhengzhou University\hspace{1.5cm} \\
\href{https://metadreamer3d.github.io/}{Project page: https://metadreamer3d.github.io/}
}


\twocolumn[{
\renewcommand\twocolumn[1][]{#1}
\maketitle
\begin{center}
    \captionsetup{type=figure}
    \includegraphics[width=1.\textwidth]{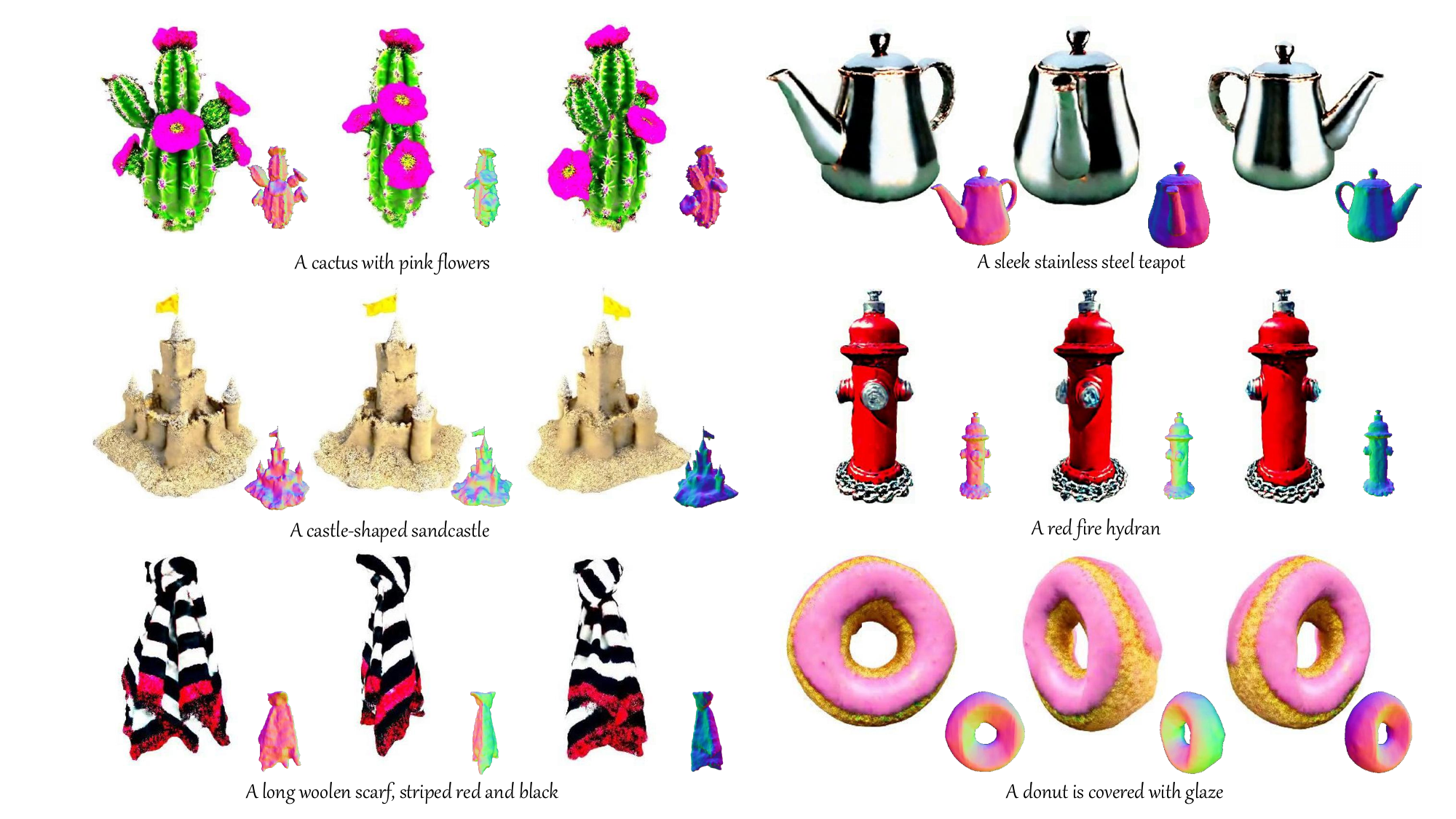}
    \captionof{figure}{\textbf{MetaDreamer for text-to-3D generation:} MetaDreamer can rapidly (20 minutes) generates high-quality 3D content based on input text. The resulting 3D objects exhibit strong multi-view consistency (no multi-headed problem) and possess complete geometry along with high-quality textures. Visit \href{https://metadreamer3d.github.io/}{https://metadreamer3d.github.io/} for an immersive visualization.}
    \label{fig:metadreamer_showcase}
\end{center}
}]

\renewcommand{\thefootnote}{\fnsymbol{footnote}} 
\footnotetext[1]{Equal contribution.} 
\footnotetext[2]{Corresponding author.} 


\begin{abstract}
Generative models for 3D object synthesis have seen significant advancements with the incorporation of prior knowledge distilled from 2D diffusion models. Nevertheless, challenges persist in the form of multi-view geometric inconsistencies and slow generation speeds within the existing 3D synthesis frameworks. This can be attributed to two factors: firstly, the deficiency of abundant geometric a priori knowledge in optimization, and secondly, the entanglement issue between geometry and texture in conventional 3D generation methods.In response, we introduce MetaDreammer, a two-stage optimization approach that leverages rich 2D and 3D prior knowledge. In the first stage, our emphasis is on optimizing the geometric representation to ensure multi-view consistency and accuracy of 3D objects. In the second stage, we concentrate on fine-tuning the geometry and optimizing the texture, thereby achieving a more refined 3D object. Through leveraging 2D and 3D prior knowledge in two stages, respectively, we effectively mitigate the interdependence between geometry and texture.
MetaDreamer establishes clear optimization objectives for each stage, resulting in significant time savings in the 3D generation process. Ultimately, MetaDreamer can generate high-quality 3D objects based on textual prompts within \textbf{20 minutes}, and to the best of our knowledge, it is the most efficient text-to-3D generation method. Furthermore, we introduce image control into the process, enhancing the controllability of 3D generation. Extensive empirical evidence confirms that our method is not only highly efficient but also achieves a quality level that is at the forefront of current state-of-the-art 3D generation techniques.  Project page at~\href{https://metadreamer3d.github.io/}{https://metadreamer3d.github.io/}. 
\end{abstract}



\section{Introduction}
The demand for 3D assets, particularly in applications such as gaming and virtual reality, is steadily increasing. However, in contrast to 2D assets, the acquisition of 3D data is notably challenging, resulting in a scarcity of such data. In order to address this issue, recent attention has been directed towards 3D generation techniques. These approaches endeavor to generate 3D assets from images or textual descriptions, offering a potential solution to the problem of 3D asset scarcity.

In the early days of 3D generation, the predominant paradigm revolved around multi-view 3D reconstruction\cite{guo2022neural,NeRF}. The fundamental idea was to gather information from diverse angles to craft a comprehensive 3D representation. However, with the advent of robust 2D models like Diffusion model\cite{sd}, a wave of innovative 3D generation methods has emerged. Broadly, these methods can be classified into two categories: text-driven \cite{dreamfusion,prolificdreamer} and single-image-driven\cite{liu2023zero,make-it-3d} 3D generation.

In the text-driven 3D paradigm, 3D content generation is guided by textual descriptions. These novel approaches\cite{dreamfusion,lin2023magic3d} utilize the natural language to create 3D representations. Text-based 3D generation methods primarily distill prior knowledge from pre-trained multimodal text-to-image generation models\cite{sd}. Their main objective is to leverage textual descriptions to generate 3D content, bridging the semantic gap between language and visual representations. While image-driven method aims to generate or reconstruct 3D structures from a single image. Single-image-based 3D generation methods incorporate 3D prior knowledge into image-based 2D diffusion models. These techniques focus on inferring 3D structures from a single image, effectively addressing the challenge of reconstructing 3D scenes from limited viewpoint information. One representative work is Zero-1-to-3\cite{liu2023zero}, which learns 3D prior knowledge from view-dependent diffusion models.  

While both image-to-3D and text-to-3D methods have shown promising results, they continue to face several challenges. Firstly, these methods are time-consuming. It takes several hours of continuous iterative optimization to generate a 3D object, consuming not only time but also a significant amount of computational resources. Another significant challenge lies in striking a balance between geometric and textural requirements. Methods based on distilling geometric priors, such as Zero123\cite{liu2023zero} and Make-it-3D\cite{make-it-3d}, excel in capturing precise geometric shapes but may fall short in delivering high-quality textures. Conversely, approaches based on 2D prior, such as\cite{dreamfields,dreamfusion,prolificdreamer,lin2023magic3d} can excel in reproducing textures but may struggle with geometric accuracy, sometimes leading to the notorious "multi-face problem". These challenges highlight the ongoing pursuit of more efficient and balanced techniques for 3D generation. 
Magic123\cite{magic123} utilizes two priors simultaneously but faces another problem that geometric and textures become entangled, resulting in training instability and failing to address the aforementioned problem of geometric texture imbalance.

We find that the fundamental cause of the aforementioned issue lies in the failure to strike a balance between geometric and texture aspects. Consequently, we propose MetaDreamer, an efficient generative 3D method that relies on the disentangling of geometric and texture priors. To the best of our knowledge, we are the first to achieve equilibrium in learning between geometry and texture through the incorporation of two distinct prior knowledge sources. As shown in Fig \ref{fig:metadreamer_showcase}, the 3D objects generated by MetaDreamer simultaneously consider both geometry and texture. In terms of geometry, the generated 3D content demonstrates strong multi-view consistency and possesses complete geometry. Regarding texture, the 3D content exhibits rich and intricate textures. Our contributions can be summarized as follows:

\begin{itemize}

    
    
    \item We introduce MetaDreamer, a novel text-to-3D generation method that employs a two-stage optimization process, from coarse to fine, to rapidly generate high-quality 3D geometry and textures.

    \item We propose using 2D and 3D prior knowledge to faithfully generate 3D content from arbitrary text prompts. In the first stage, we solely leverage 3D prior knowledge, and in the second stage, we exclusively utilize 2D prior knowledge. This approach effectively prevents the entanglement of geometric and texture priors.

    \item MetaDreamer can generate high-quality 3D content in 20 minutes. Through extensive qualitative and quantitative comparisons, we found that our method outperforms the state-of-the-art in both efficiency and quality.

\end{itemize}

\label{sec:intro}

\section{Related work}

\begin{figure*}[h!]
    \begin{minipage}{1.0\linewidth}
		\centering
		\includegraphics[width=1\linewidth]{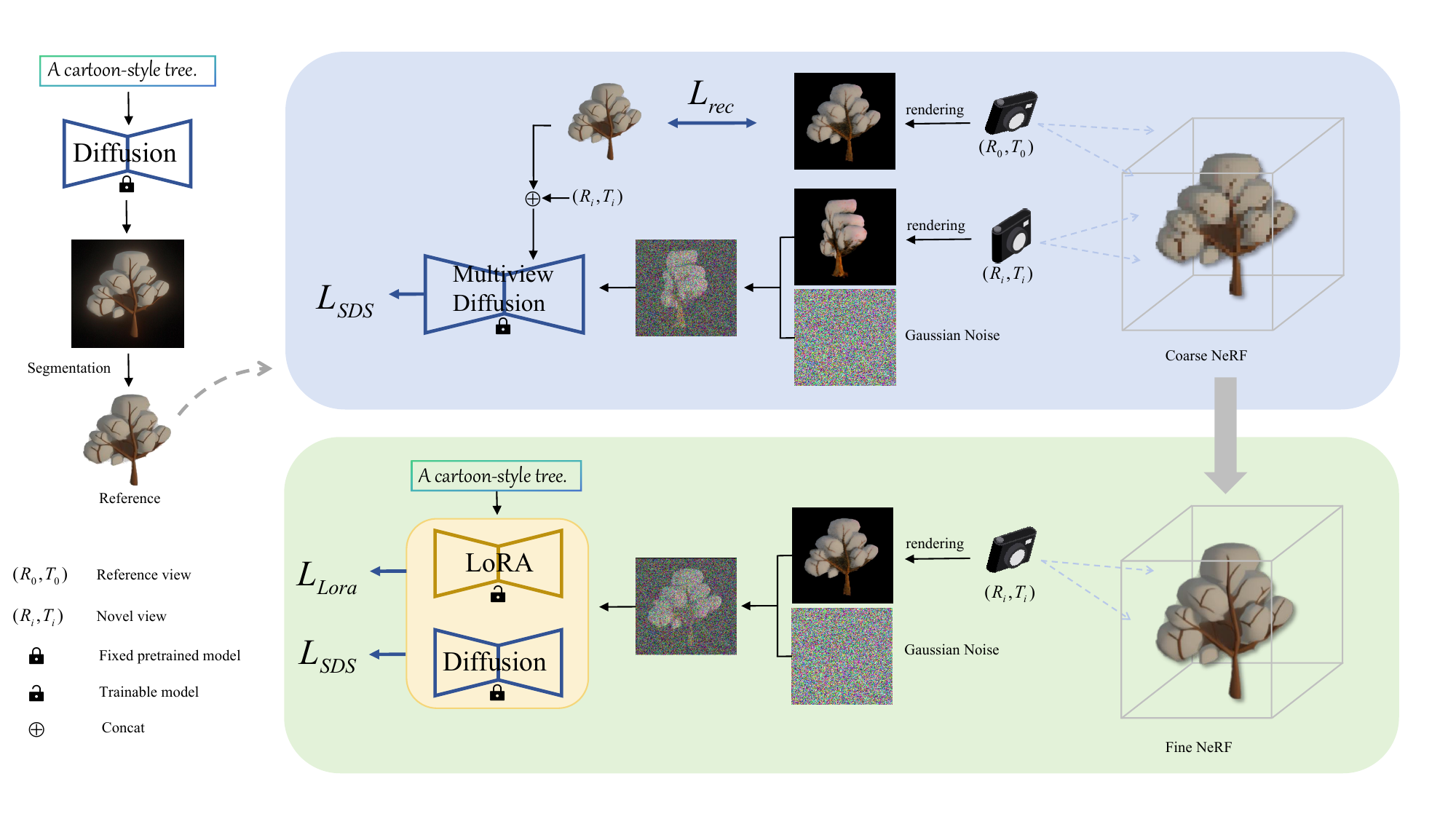}
    \end{minipage}
    \caption{
    MetaDreamer is a two-stage coarse-to-fine optimization pipeline designed to generate 3D content from arbitrary input text. In the first stage, we optimize a rough 3D model Instant-NGP\cite{NGP} guiding by a reference image and view-dependent diffusion prior model simultaneously. In the second stage, we continue to refine Instant-NGP using a text-to-image 2D diffusion prior model\cite{sd}. The entire process takes 20 minutes. The entire optimization process only takes 20 minutes.
}
    \label{fig:pipeline}
\end{figure*}

\label{sec:formatting}

\subsection{3D Reconstruction From Signal view}
Before the advent of CLIP\cite{clip} and the widespread availability of large-scale 2D diffusion models\cite{sd}, researchers frequently relied on learning 3D priors from either synthetic 3D data, as demonstrated in works such as \cite{shapenet}, or real-world scans as mentioned in \cite{common}. The representation of 3D data comes in diverse formats, encompassing 3D voxels \cite{girdhar2016learning, xie2019pix2vox}, point clouds \cite{pc1, pc2}, polygon meshes \cite{mesh1, mesh2}, and parametric models \cite{pavlakos2019expressive, zuffi2018lions, zuffi20173d}.

Recently, there has been an increasing number of works on learning to generate a 3D implicit field from a single image \cite{xu2019disn,muller2022autorf} and multiview \cite{wu2023multiview}. 
Some works leverage 2D diffusion models to enable the generation of 3D models from a single image. NeuralLift-360 \cite{NeuralLift-360} lift an in-the-wild 2D photo into a 3D object by learning probabilistic-driven 3D lifting with CLIP-guided diffusion priors and mitigates the depth errors by a scale-invariant depth ranking loss. 
A recent work Zero123 \cite{liu2023zero} finetunes the Stable Diffusion model \cite{sd} to generate a novel view of the input image based on relative camera pose. It uses fractional distillation method SDS\cite{dreamfusion} to reconstruct 3D model through distilling geometric priors of angular dependent diffusion models. 


\subsection{Text-to-3D Generation}
Recently, text-to-3D generation has become increasingly popular.
Recent advances include CLIP\cite{clip_forge}, CLIP-mesh\cite{clip_mesh}, Latent-NeRF\cite{latent-NeRF}, Dream Field\cite{dreamfields}, Score-Jacobian-Chaining\cite{sjc},  DreamFusion\cite{dreamfusion}. In CLIP-forge\cite{clip_forge}, the model is trained for shapes conditioned on CLIP text embeddings from rendered images. During inference, the embedding is provided for the generative model to synthesize new shapes based on the text. CLIP-mesh\cite{clip_mesh} and Dream Field optimized the underlying 3D representation with the CLIP-based loss. 
\my{Dreamfusion~\cite{dreamfusion} first introduce Score Distillation Sampling (SDS) that applies a pretrained diffusion to opitimize a neural radiance field, which is widely used in the following works such as~\cite{fantasia3d,dreamavatar,latent-NeRF,lin2023magic3d}.
Magic3D~\cite{lin2023magic3d} adds a finetuning phase with a textured-mesh model\cite{textual_inversion}, allowing high resolutions. ProlificDreamer~\cite{prolificdreamer} further
proposes Variational Score Distillation (VSD) that improves the diversity and details of the generated models. However, these methods only take advantage of the 2D prior in the pretrained diffusion model. The lack of 3D geometry}

Recent work, dreamfusion\cite{dreamfusion} and prolificdreamer\cite{prolificdreamer}, optimises the 3D representation of NeRF\cite{NeRF} by learning the prior knowledge of a large scale multimodal pre-trained generative model SD\cite{sd}, but they share a common problem: they only use 2D prior knowledge but lacks 3D prior knowledge, resulting in flat or even distorted object shapes. 

\section{Preliminary}
\subsection{Neural Rendering Of 3D model}
NeRF\cite{NeRF} is a technique for neural inverse rendering that consists of a volumetric raytracer and a multilayer perceptron (MLP). Rendering an image from a NeRF is done by casting a ray for each pixel from a camera's center of projection through the pixel's location in the image plane and out into the world. Sampled 3D points $\mathbf{\mu}$ along each ray are then passed through an MLP, which produces 4 scalar values as output: a volumetric density $\tau$ (how opaque the scene geometry at that 3D coordinate is) and an RGB color $c$. These densities and colors are then alpha-composited from the back of the ray towards the camera, producing the final rendered RGB value for the pixel:
\begin{equation}
\begin{aligned}
C &= \sum_i w_i c_i, \\
w_i &= \alpha_i \prod_{j<i} (1 - \alpha_j), \\
\alpha_i &= 1 - \exp(-\tau_i \|\mu_i - \mu_{i+1}\|).
\end{aligned}
\end{equation}
In the traditional NeRF use-case, we are given a dataset of input images and associated camera positions, and the NeRF MLP is trained from random initialization using a mean squared error loss function between each pixel's rendered color and the corresponding ground-truth color from the input image. This yields a 3D model (parameterized by the weights of the MLP) that can produce realistic renderings from previously-unseen views. Our model is built upon Instant-NGP\cite{NGP}, which is an improved version of NeRF for efficient highresolution rendering  with resolutions varying from 64 to 512. 

\subsection{Score Distillation Sampling}
SDS\cite{dreamfusion} is an optimization method by distilling pretrained diffusion models, also known as Score Jacobian Chaining (SJC)\cite{sjc}. It is widely used in text-to-3D\cite{dreamfusion} and imgae-to-3D\cite{liu2023zero} generation with great promise. The principle of SDS is as follows:

Given a distribution $p_t(x_t|c)$, the distribution of the forward diffusion at time $t$ of pretrained image-to-image or text-to-image diffusion model with the noise prediction network, and we denote $q_{\theta_t}(x_t|c)$ as the distribution at time $t$ of the forward diffusion process starting from the rendered image $g(\theta, c)$ with the camera $c$ and 3D parameter $\theta$, the probabilistic density distillation loss\cite{dreamfusion} optimizes the parameter $\theta$ by solving:
\begin{equation}
\min_{\theta \in \Theta} \mathcal{L}_{\mathrm{SDS}}(\theta) := \mathbb{E}_{t, c} \left[w(t) D_{\mathrm{KL}}\left( q_{t}^{\theta}(\boldsymbol{x}_{t} \mid c) \| p_{t}(\boldsymbol{x}_{t} \mid c) \right) \right]
\end{equation} where $t \sim \mathcal{U}(0.02, 0.98)$, $\epsilon \sim \mathcal{N}(0, I)$, $w(t)$ is a weighting function that depends on the timestep $t$, $x_t = \alpha_t g(\theta, c) + \sigma_t \epsilon$ is the state of the rendered image at the time $t$ of forward diffusion. With this method, we can utilize the prior knowledge of diffusion models to guide the optimization of 3D NeRF\cite{NeRF}.

\section{Proposed Method}










In this section, we introduce our MetaDreamer, an efficient and high-quality text-to-3D generation network.  As depicted in Figure \ref{fig:pipeline}, our method can be divided into two stages: the geometry stage and the texture stage. In the geometry stage, we obtain a coarse 3D representation, while in the texture stage, we further refine the geometry and enhance its texture. 
Through the optimization in two stages, we are able to disentangle the interaction between geometry and texture during the optimization process. This makes the optimization objectives for each stage more explicit, which is crucial for improving both the efficiency and quality of 3D generation.

\subsection{Preparatory work}
MetaDreamer takes a textual prompt as input and generates a 3D model as output.
In the first stage, we employ text-to-image diffusion model\cite{sd} to generate 2D reference images $\mathbf{I}_r$ for guiding geometric learning. We also leverage an off-the-shelf segmentation model, SAM\cite{sam}, to segment the foreground. The extracted mask, denoted as $\mathbf{M}$ is a binary segmentation mask and will be used in the optimization. To prevent flat geometry collapse, i.e. the model generates textures that only appear on the surface without capturing the actual geometric details, we further extract the depth map from the reference view by the pretrained MiDaS\cite{depth}. The foreground image is used as the input, while the mask and the depth map are used in the optimization as regularization priors. 

\subsection{Geometric optimization}
During the geometric optimization stage, MetaDreamer acquires knowledge from the fusion of reference images and pretrained geometric prior model\cite{liu2023zero}. In this stage, we focus on learning the overall geometric shape, and care less about geometric details and textures.
The objective is to rapidly establish the fundamental geometric structure of the 3D object. In terms of 3D representation, we employ the implicit parameterization model NeRF\cite{NeRF}. NeRF excels in capturing complex geometric properties, making it the ideal choice for our goal of swiftly acquiring the geometric representation from reference images. We use the pretrained multi-view diffusion model and reference image priors separately to guide the learning of 3D NeRF.
\paragraph{View-dependent Diffusion Prior}
The pretrained view-dependent prior diffusion model zero123xl\cite{liu2023zero} is used to guide the optimization in our Method. It it fine-tuned from an image-to-image diffusion model using the Objaversexl\cite{objaverse} dataset, the largest open-source 3D dataset that consists of 10 million models. Given the diffusion model denoiser $\theta$, the diffusion time step $t \sim [1, 1000]$, the embedding of the input view and relative camera extrinsics $c(x, R, T)$, the view-dependent diffusion model is optimized by the following constraints:
\begin{equation}
    \min_{\theta} \mathbb{E}_{z \sim \mathcal{E}(x), t, \epsilon \sim \mathcal{N}(0,1)} \left\| \epsilon - {\epsilon}_{\theta}(z_t, t, c(x, R, T)) \right\|_2^2
\end{equation} where $\epsilon \sim \mathcal{N}(0, I)$, $z_t = \alpha_t x_{R,T} + \sigma_t \epsilon$ is the target image with noise. In this way, a view-dependent 3D prior diffusion model ${\epsilon}_{\theta}$ can be obtained.

\paragraph{Geometry Score Distillation Sampling}
In this process, we first randomly initialized a 3D model ${\epsilon}_{\theta}$ with parameter $\theta \in \Theta$, where $\Theta$ is the space of $\theta$ with the Euclidean metric. Then we randomly sample a position and angle for a ray in a 3D scene, with the ray's position and direction represented in spherical coordinates as $\mathbf{r}=(\rho , \vartheta , \varphi)$, and render the shaded NeRF model at $256 * 256$ resolution  ${g}(\theta,\mathbf{r},c)$. After that, we perform the forward diffusion process: add random Gaussian noise to the rendered image. The hidden layer noise image at step $t$ is represented as $x_t = \alpha_t g(\theta, \mathbf{r}, c) + \sigma_t \epsilon$.

We then make direct use of the loss function of the diffusion model: a noise estimate is made on the noise graph and the MSE loss is used to constrain it:
\begin{equation}
\begin{aligned}
\mathcal{L}_{3 D}=\mathbb{E}_{t, \epsilon}\left[w(t)\|\left({\epsilon}_{\text{pretrain1}}\left(\mathbf{x}_{t} ; \mathbf{I}^{r}, t, c\right)-\epsilon\right)\|_{2}^{2}\right] \\
\end{aligned}
\end{equation} where $c$ is the camera poses passed to view-dependent diffusion model. Intuitively, Geometry-based SDS leverages the multi-view geometric relationships of the view-dependent diffusion model to encourage 3D consistency. It's important to note that during this process, the diffusion model parameters are frozen.


\paragraph{Reference view Prior}
The reference image prior plays a crucial role in ensuring the 3D fidelity. $L_{\text{rec}}$ is imposed in the geometry stage as one of the major loss functions to ensure the rendered image from the reference viewpoint ($v_r$, assumed to be front view) is as close to the reference image $\mathbf{I}_r$ as possible. We adopt the mean squared error (MSE) loss on both the reference image and its mask as follows:
\begin{equation}
\begin{aligned}
\mathcal{L}_{\text {rec }} = \lambda_{\text{rgb}} \|\mathbf{M} \odot\left(\mathbf{I}^{r}-g\left({\theta},\mathbf{v}^{r}\right)\right) \|_2^2 \\
+ \lambda_{\text{mask}} \| \mathbf{M} - M (g({\theta},\mathbf{v}^r)) \|_2^2
\end{aligned}
\end{equation}
where $\theta$ is the NeRF parameters to be optimized, $\mathbf{M}$ is a binary segmentation mask of $\mathbf{I}_r$, $\odot$ is the Hadamard product, $g(\mathbf{\theta},\mathbf{v}_r,c)$ is the NeRF rendered view from the viewpoint $\mathbf{v}_r$, $M(\cdot)$ is the foreground mask acquired by integrating the volume density along the ray of each pixel. $\lambda_{\text{rgb}}$ and $\lambda_{\text{mask}}$ are the weights for the foreground RGB and the mask.

\paragraph{Depth Prior}
The depth prior is employed to prevent excessively flat or concave 3D representations. Relying solely on appearance reconstruction losses can lead to suboptimal geometric results, given the inherent ambiguity of reconstructing 3D content from 2D images. This ambiguity arises because the 3D content could exist at various distances while still appearing as the same 2D image, potentially resulting in flat or concave geometries, as observed in prior research (NeuralLift-360\cite{NeuralLift-360}). To alleviate this problem, we introduce depth regularization. We utilize a pretrained monocular depth estimator\cite{depth} to obtain the pseudo depth ($d_r$) for the reference image. The NeRF model's depth output ($d$) from the reference viewpoint should closely align with the depth prior. However, due to disparities between the two sources of depth estimation, using the Mean Squared Error (MSE) loss is not ideal. Instead, we employ the normalized negative Pearson correlation as the depth regularization term.
\begin{equation}
\mathcal{L}_{d}=\frac{1}{2}\left[1-\frac{\operatorname{Cov}\left(\mathbf{M} \odot d^{r}, \mathbf{M} \odot d\right)}{\operatorname{Cov} \left(\mathbf{M} \odot d^{r}\right) \operatorname{Var}(\mathbf{M} \odot d)}\right]
\end{equation} where $\text{Cov}(\cdot)$ denotes covariance and $\text{Var}(\cdot)$ measures standard deviation.

\paragraph{Geometry regularizers}
One of the NeRF limitations is the tendency to produce high-frequency artifacts on the surface of the object. To address this, we enforce the smoothness of the normal maps of geometry for the generated 3D model following \cite{Realfusion}. We use the finite differences of the depth to estimate the normal vector of each point, render a 2D normal map $n$ from the normal vector, and impose a loss as follows:
\begin{equation}
    \mathcal{L}_{n}=\|\mathbf{n}-\tau(g(\mathbf{n}, k))\|
\end{equation} where $\tau(\cdot)$ denotes the stopgradient operation, and $g(\cdot)$ is a Gaussian blur. The kernel size of the blurring, $k$, is set to 9 $\times$ 9.

\subsection{Texture optimization}
In texture modeling stage, MetaDreamer primarily focuses on further refining the coarse geometric model obtained in the first stage, encompassing both geometry and textures. Similar to the first stage, in this stage, we heavily rely on pretrained text-to-image diffusion models ${\epsilon}_{\phi}$. We transfer the prior knowledge from these 2D images into the 3D model through SDS\cite{dreamfusion}.  It's worth noting that there are domain gap between 2D and 3D. To narrow this domain gap, we employ an efficient parameter fine-tuning method, Lora\cite{lora} to fine-tune the diffusion model.

\paragraph{Texture Score Distillation Sampling}
Given a text-to-image diffusion prior model $\epsilon_{\text{sd}}$, and a coarse geometric model $g(\theta)$ (obtained in the geometric stage), we employ SDS to further refine the geometric textures. Specifically, we first encode the rendered view $g(\theta, c)$  as latent $z_0$, $z_t = \alpha_t z_0 + \sigma_t \epsilon$  is the noisy representation of the lantent $z_0$ after $t$ steps of forward diffusion and adds noise to it, and guesses the clean novel view guided by the input text prompt. Roughly speaking, SDS translates the rendered view into an image that respects both the content from the rendered view and the prompt. The texture score distillation sampling loss is as follows:
\begin{equation}
\begin{aligned}
\mathcal{L}_{2 D}(\theta)=\mathbb{E}_{t, \epsilon}\left[w(t)\|\left({\epsilon}_{\text{sd}}\left(\mathbf{z}_{t} ; t, c\right)-\epsilon\right)\|_{2}^{2}\right] \\
\end{aligned}
\end{equation} where {$c$} represents the camera's intrinsic parameters, $\theta$ is learnable parameter of NeRF. It is worth noting that the parameters of the stable diffusion  model $\epsilon_{sd}$ are frozen.

\paragraph{2D-to-3D Domain Adaptation}
Despite the powerful prior knowledge in diffusion models, applying this prior knowledge directly to guide 3D generation is not ideal due to the gap between the 2D and 3D domains. To solve this problem, we employ Lora\cite{lora} to fine-tune the diffusion models since it's great capacity for few-shot fine-tuning. The training loss for Lora is as follows:
\begin{equation}
\begin{aligned}
\mathcal{L}_{Lora}(\theta,\phi)=\mathbb{E}_{t, \epsilon}\left[w(t)\|\left({\epsilon}_{\phi}\left(\mathbf{z}_{t} ; t, c\right)-\epsilon\right)\|_{2}^{2}\right] \\
\end{aligned}
\end{equation} where $\epsilon_{\phi}$ is the small learnable Unet\cite{unet} condition by camera parameter c and time embedding.

\paragraph{Opacity regularization}
To prevent high-frequency artifacts in 3D space, we first introduce a novel opacity regularization technique.
In single-object 3D generation, this penalty term plays a crucial role in accelerating convergence and improving geometric quality: it significantly suppresses unnecessary blank filling and mitigates noise diffusion:
\begin{equation}
\begin{aligned}
\mathcal{L}_{\text {reg}}= \sum_{i} \sum_{j} \left \| w_{ij} \right \| ^2 \\
\text { s.t. } w_{ij} \notin \mathbf{C}_{max} \\
\end{aligned}
\end{equation} where $w_{ij}$ is the rendering weight, and $\mathbf{C}_{max}$ is the largest connected component of the rendering weight matrix.

\begin{figure*}[]
    \begin{minipage}{1.2\linewidth}
		\centering
		\includegraphics[width=1\linewidth]{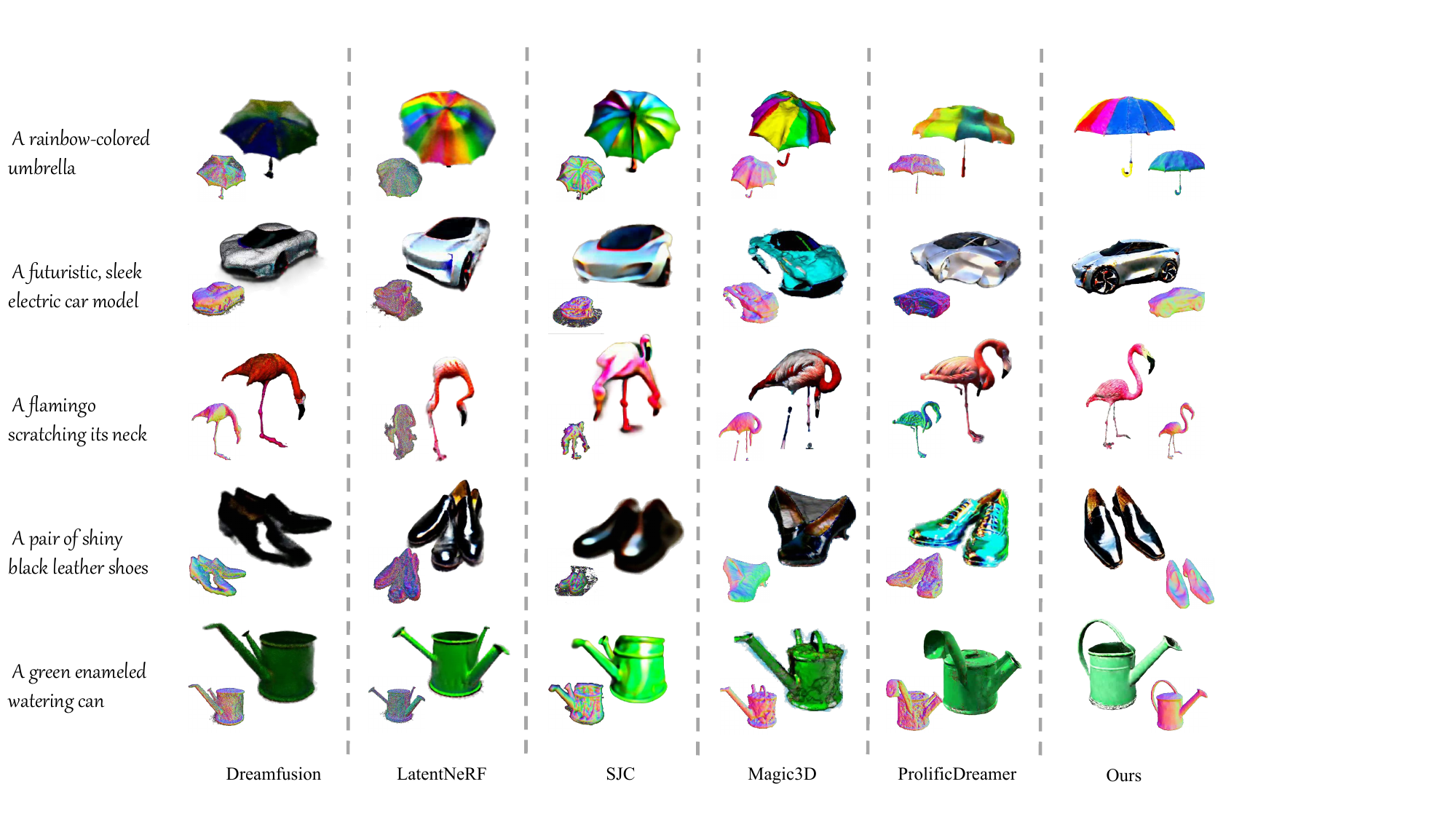}
    \end{minipage}
    \caption{Text-to-3D samples generated by MetaDreamer from scratch. Our base model is Stable Diffusion and we do not employ any other assistant model or user-provided shape guidance (see Table 1). See our accompanying videos in our project page for better visual quality.
}
    \label{fig:comparison}
\end{figure*}

\section{Experiments}

\subsection{Qualitative Analysis}
We qualitatively compare MetaDreamer with other advanced 3D methods: Dreamfusion\cite{dreamfusion}, LatentNeRF\cite{latent-NeRF}, SJC\cite{sjc}, Magic3D\cite{lin2023magic3d}, and ProlificDreamer\cite{prolificdreamer}. As seen in Figure \ref{fig:comparison}, Other methods, only guided by 2D priors such as Dreamfusion\cite{dreamfusion}, Magic3D\cite{lin2023magic3d}, etc., shares a common issue: the Janus problem (also known as the multi-headed problem). Moreover, their geometry is incomplete, not smooth, and contains numerous holes. We attribute these problems to their failure to introduce 3D priors. In comparison, our method solves the multi-head problem well and has a more complete and smooth 3D normal. As for texture, despite our model requiring only 20 minutes of optimization, its textures are remarkably detailed, comparable to or even surpassing current state-of-the-art methods. This improvement in texture quality is attributed to our geometry-texture decoupled optimization approach.

\subsection{Quantitative comparison}

\paragraph{2D metrics}
In the context of text-based 3D, where there are no standardized metrics for 3D objects, we employ 2D metrics for evaluation. We evaluate on three variants of CLIP\cite{clip}: CLIP B/32, CLIP B/16, and CLIP B/14. Specifically, we indirectly measure the similarities of CLIP  between text prompts and 3D objects by comparing the similarity of 2D renderings of text prompts and 3D objects. We compare our method with state-of-the-art text-to-3D methods, such as DreamFusion\cite{dreamfusion}, ProlificDreamer\cite{prolificdreamer}, SJC\cite{sjc},Latent-NeRF\cite{latent-NeRF} and Magic3D\cite{lin2023magic3d}. Additionally, we assess the similarity between 2D images generated by our diffusion model\cite{sd} and the corresponding text, denoted as GT. In theory, when evaluating 3D quality with this method, the similarity cannot exceed this value. Among all methods, MetaDreamer obtains the highest CLIP similarity score, closest to the GT score. This indirectly demonstrates its ability to better maintain consistency between 3D objects and input text. 

\paragraph{3D metrics}
$\text{T}^{3}$Bench~\cite{he2023t} privodes a comprehensive text-to-3D benchmark to assess the quality of the generated 3D models. They introduce two metrics:
the quality score and the alignment score. The quality metric utilizes multi-view text-image scores and regional convolution to evaluate the visual quality and view inconsistency. The alignment metric relies on multi-view captioning and Large Language Model (LLM) evaluation to measure text-3D consistency. Using the generic prompts they provided, we conduct comparative experiments under the setting of $Single Object$ where only one single object and its description are mentioned each prompt. Our methond achieves the highest scores on both quality metric and alignment metric as shown in Table~\ref{T3bench}.

\begin{table}[]
\centering
\scalebox{0.95}
{
\begin{tabular}{cccc}
\hline
Mothod          & CLIP B/32$\uparrow$ & CLIP B/16$\uparrow$ & CLIP L/14$\uparrow$ \\ \hline
GT              & 0.2947   & 0.2913    & 0.2715    \\
DreamFusion     & 0.2415    & 0.2303    & 0.2432    \\
LatentNeRF      & 0.2373    & 0.2301    & 0.2210    \\
SJC             & 0.2211    & 0.2365    & 0.2313    \\
Magic3D         & 0.2673    & 0.2701    & 0.2610    \\
ProlificDreamer & 0.2715    & 0.2829    & 0.2669    \\ \hline
Ours            & \textbf{0.2869}    & \textbf{0.2900}    & \textbf{0.2710}    \\ \hline
\end{tabular}
}
\caption{The consistency of the 3D model with the provided text prompt is assessed by computing the similarity between multiple randomly rendered views of the 3D model and the given text prompt. GT represents the 2D image generated by the text-to-image Diffusion model\cite{sd}
}
\end{table}

\begin{table}[]
\begin{tabular}{cccc}
\hline
Mothod & Quakity$\uparrow$ & Alignment$\uparrow$ & Average$\uparrow$ \\ \hline
DreamFusion & 24.9 & 24.0& 24.4  \\
LatentNeRF & 34.2 & 32.0& 33.1  \\
SJC & 26.3 & 23.0& 24.7  \\
Magic3D & 38.7  & 35.3 & 37.0  \\ 
ProlificDreamer& 51.1 & 47.8 & 49.3 \\
\hline
Ours& \textbf{54.6}& \textbf{55.8} & \textbf{55.2} \\ 
\hline
\end{tabular}
\caption{Comparisons in terms of $\text{T}^{3}$Bench benchmarks.
}
\label{T3bench}
\end{table}

\subsection{Efficiency Evaluation}
To demonstrate the efficiency of MetaDreamer, we compare it with popular text-to-3D generation methods in terms of training iteration counts and time consumption. Table\ref{costtime} reveals that the average number of iterations for mainstream methods currently stands at 26,000, with an average duration of 2.5 hours. In contrast, while MetaDreamer requires only 1,300 iterations (including 300 iterations in the first stage and 1,000 iterations in the second stage), it can generate 3D models comparable to or even better than mainstream methods. The entire process takes only 20 minutes, saving 2 hours compared to mainstream methods. This efficiency improvement is attributed to our disentangling training of geometry and texture.

\begin{figure*}[ht]
    \begin{minipage}{1.\linewidth}
		\centering
		\includegraphics[width=1\linewidth]{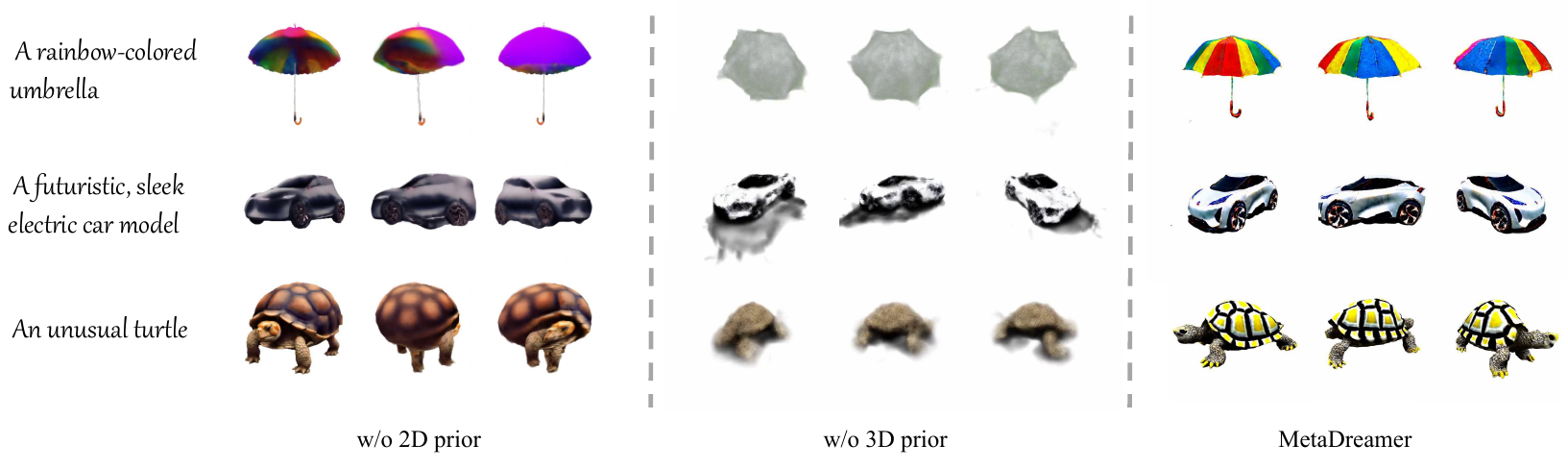}
    \end{minipage}
    \caption{Qualitative comparison: The left side is the multi-view rendering image at the coarse stage, and the right side is the multi-view rendering image at the refined stage.
}
    \label{ablation}
\end{figure*}

\begin{table*}[ht]
\centering
\scalebox{1.0}{
\begin{tabular}{ccccccc}
\hline
Method    & LatenNeRF & DremFusion & Magic3D & SJC & ProlificDreamer & MetaDreamer \\ \hline
Time(min) & 100        & 60         & 125     & 65          & 420             & 20          \\
Iterations      & 20000     & 10000      & 20000   & 10000      & 70000           & 1300        \\ \hline
\end{tabular}
}
\caption{Comparison of average training times between MetaDreamer and various text-based 3D methods. All experiments were conducted on a single NVIDIA A100 GPU. All experimental settings (number of iterations, random seeds, etc.) followed the official default settings of threestudio\cite{threestudio2023}.}
\label{costtime}
\end{table*}

\subsection{Ablation Study}
In this section, we qualitatively analyze the effects of different prior knowledge on MetaDreamer. Specifically, we conduct two experiments: using only 3D priors and using only 2D priors. From the Fig \ref{ablation}, it is apparent that when solely utilizing 3D prior knowledge (optimized for 300 iterations in the first stage), we obtain a rough geometric model demonstrating good geometric integrity and viewpoint consistency. However, it still lacks geometric details and clear textures. Conversely, when exclusively employing 2D prior knowledge (optimized for 1000 iterations in the second stage), we only obtain a very blurry residue, which we attribute to the lack of 3D prior knowledge causing the 3D object not to converge. When combining 2D and 3D prior knowledge in a two-stage manner, we achieve a perfect 3D object. It is evident that the geometric and texture details missed in the first stage are compensated. Experimental results demonstrate the complementary nature of the two-stage optimization: the coarse model from the first stage aids in accelerating geometric convergence in the second stage, while the diffusion model and strong semantic and 2D priors in the second stage contribute more imaginative power, helping to compensate for the geometric and texture deficiencies from the first stage.














\section{Conclusion}
In this work, we have proposed MetaDreamer, an efficient and high-quality text-to-3D generation method. Our approach leverages two different types of prior knowledge: geometric priors (3D) and texture priors (2D), and adapts the domain gap between 2D and 3D knowledge using efficient parameter fine-tuning method, LoRA. To prevent the entanglement of the two types of priors, we use only geometry priors in the coarse stage and only texture priors in the fine stage. Our MetaDreamer can generate high-quality 3D content within 20 minutes. Abundant qualitative and quantitative comparative experiments demonstrate that our method surpasses the state-of-the-art level in both efficiency and quality.

\section{Future Work}
MetaDreamer performs at the state-of-the-art level in terms of both efficiency and quality, but it still has some limitations. For example, it performs poorly in multi-object generation tasks due to the lack of prior knowledge about multiple objects in geometric priors. We have attempted to introduce multi-object priors using powerful multimodal text-image pretraining models, but the results have not been ideal, and they come with significant time consumption. Therefore, we will address this challenge  in the next stage of our work by injecting more multi-object geometric prior knowledge into the model.


{\small
\bibliographystyle{ieee_fullname}
\bibliography{egbib}
}

\end{document}